\documentclass[sigconf]{acmart}
\usepackage{amsmath}
\usepackage{amsfonts}
\usepackage{algorithmic}
\usepackage{graphicx}
\usepackage{textcomp}
\usepackage{xcolor}
\usepackage{hyperref}
\usepackage{subfigure}
\usepackage{multirow}
\usepackage{booktabs} 
\usepackage{bm}
\usepackage{subcaption}
\usepackage{pifont}
\usepackage{xcolor}
\usepackage{tabularx}
\usepackage{float}
\usepackage{threeparttable}

\AtBeginDocument{%
  }

\setcopyright{acmlicensed}
\copyrightyear{2018}
\acmYear{2018}
\acmDOI{XXXXXXX.XXXXXXX}

\acmConference[Conference acronym 'XX]{Make sure to enter the correct
  conference title from your rights confirmation emai}{June 03--05,
  2018}{Woodstock, NY}
\acmISBN{978-1-4503-XXXX-X/18/06}

\begin{document}
\settopmatter{printacmref=false} % Removes citation information below abstract
\renewcommand\footnotetextcopyrightpermission[1]{} % Removes copyright footnote
\pagestyle{plain} % Removes running headers

\title{Exploring Gaze Pattern Differences Between Autistic and Neurotypical Children: Clustering, Visualisation, and Prediction}


\author{Weiyan Shi}
\affiliation{%
  \institution{Singapore University of Technology and Design}
  \country{Singapore}
}

\author{Haihong Zhang}
\affiliation{%
  \institution{A*STAR Institute for Infocomm Research}
  \country{Singapore}
}

\author{Wei Wang}
\affiliation{%
  \institution{University of Wisconsin}
  \city{Madison}
  \state{Wisconsin}
  \country{United States}
}

\author{Kenny Tsu Wei Choo}
\affiliation{%
  \institution{Singapore University of Technology and Design}
  \country{Singapore}
}
\authornote{Corresponding author}

\renewcommand{\shortauthors}{Weiyan Shi et al.}

\begin{abstract}
Autism Spectrum Disorder (ASD) affects children's social and communication abilities, with eye-tracking widely used to identify atypical gaze patterns. While unsupervised clustering can automate the creation of areas of interest for gaze feature extraction, the use of internal cluster validity indices, like Silhouette Coefficient, to distinguish gaze pattern differences between ASD and typically developing (TD) children remains underexplored.
We explore whether internal cluster validity indices can distinguish ASD from TD children. Specifically, we apply seven clustering algorithms to gaze points and extract 63 internal cluster validity indices to reveal correlations with ASD diagnosis. Using these indices, we train predictive models for ASD diagnosis. Experiments on three datasets demonstrate high predictive accuracy (81\% AUC), validating the effectiveness of these indices.
\end{abstract}





\maketitle

\section{INTRODUCTION}

Autism Spectrum Disorder (ASD) is a developmental disorder marked by challenges in social interaction, communication, and repetitive behaviors, and is increasingly prevalent worldwide~\cite{world2019autism, neik2014prevalence}. Thus, diagnosis and intervention tools are highly demanded. Currently, diagnostic tools focus on eye-tracking as it can capture atypical gaze patterns, which are considered a key indicator of ASD~\cite{guillon2014visual}. To be specific, extensive gaze signal data collected in eye-tracking experiments highlight how individuals with ASD focus differently on social cues compared to typically developing (TD) children. Areas of Interest (AOIs) are commonly used to analyze gaze pattern differences between ASD and TD individuals by measuring average gaze points and time on specific regions~\cite{nayar2022constellation, tsuchiya2021diagnosing}. However, most works rely on manual inefficient AOI definition~\cite{del2018investigation, bataineh2018visual, almourad2018analyzing}.

Recently, some automated methods like clustering have emerged for defining AOIs. Chang et al. applied KMeans~\cite{hartigan1979algorithm} to cluster ASD and TD children’s gaze points in a video into two regions, which were used to define AOIs. The Silhouette Coefficient (SC), a classic internal cluster validity index~\cite{rousseeuw1987silhouettes}, was employed to evaluate the quality of the generated AOIs. Additionally, they extracted gaze features such as the average gaze points and time within each AOI for further analysis~\cite{chang2021computational}. However, beyond evaluating AOI generation, a significant gap remains in exploring whether internal cluster validity indices can represent gaze patterns to differentiate ASD and TD children.

To address this gap, we conducted an exploratory study using internal cluster validity indices to analyze gaze pattern differences between ASD and TD children. Specifically, we first perform seven unsupervised clustering methods on gaze data across three datasets we collected. Subsequently, we extract nine internal cluster validity indices from the clustering results. Finally, we utilized these indices features to train an ensemble of multiple predictive machine-learning models for ASD predictions. Our contributions can be summarized as:
\begin{itemize}
    \item We collected three eye-tracking datasets comprising 8,790 entries of ASD and TD children's gaze points on structured, semi-structured, and non-structured visual stimuli.
    \item We conducted the first comprehensive significance test of clustering algorithms combined with internal cluster validity indices across three complex visual stimuli datasets to distinguish between ASD and TD children, achieving a significant proportion of internal cluster validity indices at $74.1\%$.
    \item We performed the first comprehensive machine learning benchmark test using only internal cluster validity indices features across three datasets to predict ASD, achieving an average $81\%$ AUC,  highlighting the significance of internal cluster validity indices in ASD prediction.
\end{itemize}
\section{RELATED WORK}
Individuals with ASD often exhibit impaired attention to social cues, including, but not limited to, restricted focus on eyes and faces~\cite{del2018investigation, bataineh2018visual, almourad2018analyzing}, heightened attention to non-social stimuli, increased attention to detail~\cite{kwon2019typical, almourad2018analyzing}, and challenges in shifting attention compared to TD controls~\cite{frazier2017meta}. Related attributes can be derived in gaze data using gaze points and pupil dilation~\cite{ccetintacs2023detection}, and their dynamics measurements.  Hence, a common methodology involves identifying AOI on static images~\cite{nayar2022constellation} or sequentially on dynamic videos~\cite{tsuchiya2021diagnosing}, and examining the gaze attributes in association with the AOIs. For example, ~\cite{nayar2022constellation} manually drew social and non-social AOIs and examined gaze variables such as dwell time and gaze count. While straightforward and intuitive, the existing approach may not effectively capture the distinctive structures of gaze patterns especially when the AOIs are inconsistent or ambiguous. At the same time, the approach also requires a labor-intensive manual labeling process which may become prohibitively expensive for massive gaze data sets. 

To overcome the limitations of manual AOI selection, clustering-based methods offer an automated and objective alternative for analyzing gaze patterns. Naqshbandi et al. found that both K-Means~\cite{hartigan1979algorithm} and OPTICS~\cite{ankerst1999optics} achieve higher success rates compared to the manual selection of AOIs~\cite{naqshbandi2016automatic}. Internal cluster validity indices are essential for assessing clustering quality by evaluating the compactness and separation of clusters based solely on the dataset's intrinsic properties~\cite{hassan2024cluster}. For example, Chang et al. used SC to evaluate the generation of AOIs based on clustered gaze points. However, as previously mentioned, no studies have explored whether internal cluster validity indices can effectively represent gaze patterns to differentiate between ASD and TD children.
\section{GAZE POINT DATASETS OF ASD AND TD CHILDREN ACROSS DIFFERENT TYPES OF STIMULI}
We utilize three datasets of eye-tracking in ASD children to exploit three types of visual stimuli, including structured, semi-structured, and non-structured categories, respectively. (i) The \textbf{structured} category is provided in Qiao's study~\cite{he2021automatic}, which involved 74 children (50 ASD, 24 TD) aged around 5. They used a Tobii X120 eye tracker to collect gaze point data, where the child viewed four objects placed at the top, bottom, left, and right, with cartoon faces, fingers, or arrows as directional cues. (ii) The \textbf{semi-structured} category is from Cilia's study~\cite{cilia2022eye}, which analyzed gaze patterns of 59 children (30 TD, 29 ASD) aged 3-12. They used an SMI Red-M eye tracker to collect eye-tracking data as subjects viewed photos and videos with content ranging from colorful balloons and cartoons to human presenters guiding attention. (iii) The \textbf{non-structured} category is represented by the Saliency4ASD dataset~\cite{gutierrez2021saliency4asd}, where 28 children (14 ASD, 14 TD) aged 5-12 used a Tobii T120 eye tracker to view 500 images selected from three public eye-tracking datasets, featuring diverse semantic content such as animals, buildings, and people. Qiao and Cilia only provided gaze points, while Saliency4ASD provided visual stimuli along with gaze points. For the experimental dataset, each entry represents one complete instance of a child viewing a single visual stimulus. The Qiao dataset contains 8,790 entries, Cilia's dataset has 2,612 entries, and Saliency4ASD includes 602 entries.
\section{UNSUPERVISED CLUSTERING ON GAZE POINTS ACROSS THREE DATASETS}

During preprocessing, we prepare data for clustering algorithms by removing invalid gaze points and extracting data from individual children and specific experiment segments.

For a thoughtful analysis, we employed four different types of clustering algorithms and selected the most superior ones from each type as follows: 
(i) \textbf{Partitioning Methods}: K-Means~\cite{hartigan1979algorithm} and K-Medoids~\cite{park2009simple}. These two methods employ different mechanisms to cluster data points, so more feature patterns can be captured. 
(ii) \textbf{Hierarchical Methods}: Agglomerative Clustering (AC)~\cite{murtagh2014ward} and BIRCH~\cite{zhang1996birch}. AC and BIRCH cluster data into two different structures, a hierarchy, and a tree, thus exploring the most suitable structure to present data.
(ii) \textbf{Density-Based Methods}: DBSCAN~\cite{schubert2017dbscan} and OPTICS~\cite{ankerst1999optics}. They identify clusters based on the density of points, allowing them to handle clusters of arbitrary shapes and filter out noise.
(iv) \textbf{Model-Based Methods}: GMM~\cite{dempster1977maximum}. GMM clusters data to a mixture of separate Gaussian distributions, thus improving their diversity. 

For each entry across the three datasets, we applied all seven clustering algorithms, generating seven clustering outcomes per entry. To ensure a fair comparison, all clustering methods were optimized to their best configurations using grid search and evaluated based on SC.
\section{INTERNAL CLUSTER VALIDITY INDICES EXTRACTION ACROSS THREE DATASETS}
To extract comprehensive internal cluster validity indices from the clustering results, we selected nine widely used indices. While some of these indices have been used in previous studies~\cite{chang2021computational}, but they were not applied comprehensively.
These indices can be summarized as follows: (i) \textbf{Compactness and Separation} indices include Calinski–Harabasz (CH)~\cite{calinski1974dendrite}, Davies–Bouldin’s index (DB)~\cite{davies1979cluster}, DaviesBouldin* index (DB*)~\cite{kim2005new}, Dunn’s index (DI)~\cite{dunn1973fuzzy}, and Chou-SuLai (CSL)~\cite{chou2004new}, which evaluate clustering quality by balancing intra-cluster compactness with inter-cluster separation. (ii) \textbf{Similarity and Separation} indices include SC~\cite{rousseeuw1987silhouettes}, which assesses clustering by comparing intra-cluster similarity to inter-cluster separation. (iii) \textbf{Generalized Measures} include the Generalized Dunn index (GD33)~\cite{bezdek1998some}, which considers complex cluster shapes with generalized distance metrics.  
(iv) \textbf{Comprehensive Indices} such as the PBM index (PBM)~\cite{pakhira2004validity} and STR index (STR)~\cite{starczewski2017new} combine multiple features like compactness, separation, and stability for a holistic clustering evaluation. Subsequently, we conducted a Mann-Whitney U test~\cite{mcknight2010mann} on these observation values to explore statistically significant differences between ASD and TD children groups.

Table~\ref{tab:test} shows the experimental results. KMedoids, BIRCH, and OPTICS achieved strong performance, with more than 80\% of their indices showing significance. KMedoids performed the best with 89\%. In contrast, the other algorithms were less effective at distinguishing between ASD and TD children, with significance ratios below 70\%. The Qiao dataset performed the best overall, with all algorithms showing strong results across all indices. This may be due to the fact that the gaze points in Qiao were generated from highly structured visual stimuli. In contrast, the less structured visual stimuli in the Cilia and Saliency4ASD datasets resulted in significantly lower proportions of significant indices, at 57.1\% and 68.3\%, respectively.

\begin{table*}[t]
\centering
\begin{threeparttable}
\caption{The experimental results of clustering algorithms in three datasets. We performed a Mann-Whitney U test on 10 internal cluster validity indices in Qiao, Cilia, and Saliency4ASD datasets. ($\ast\ast\ast$: $p < 0.01$, $\ast\ast$: $0.01 \leq p < 0.05$, and $\ast$: $p \geq 0.05$)}
\begin{tabular*}{\textwidth}{@{\extracolsep{\fill}}llccccccccc}
\toprule
\textbf{Algorithm} & \textbf{Dataset} & \textbf{SC} & \textbf{CH} & \textbf{DB} & \textbf{CSL} & \textbf{DI} & \textbf{DB*} & \textbf{GD33}  & \textbf{PBM} & \textbf{STR} \\
\midrule
\multirow{3}{*}{KMeans} & Qiao & $\ast\ast\ast$ & $\ast\ast\ast$ & $\ast\ast\ast$ & $\ast\ast\ast$ & $\ast\ast\ast$ & $\ast\ast\ast$ & $\ast\ast\ast$ & $\ast\ast$ & $\ast\ast\ast$ \\
& Cilia & $\ast$ & $\ast\ast\ast$ & $\ast$ & $\ast$ & $\ast$ & $\ast$ & $\ast$ & $\ast\ast\ast$ & $\ast\ast$ \\
& Saliency4ASD & $\ast\ast\ast$ & $\ast\ast\ast$ & $\ast$ & $\ast$ & $\ast\ast\ast$ & $\ast$ & $\ast\ast$ & $\ast\ast\ast$ & $\ast\ast\ast$ \\
\midrule
\multirow{3}{*}{KMedoids} & Qiao & $\ast\ast\ast$ & $\ast\ast\ast$ & $\ast\ast\ast$ & $\ast\ast\ast$ & $\ast\ast\ast$ & $\ast\ast\ast$ & $\ast\ast\ast$ & $\ast\ast$ & $\ast\ast\ast$ \\
& Cilia & $\ast\ast\ast$ & $\ast\ast\ast$ & $\ast\ast\ast$ & $\ast\ast$ & $\ast\ast\ast$ & $\ast\ast\ast$ & $\ast\ast\ast$ & $\ast\ast\ast$ & $\ast\ast\ast$ \\
& Saliency4ASD & $\ast\ast\ast$ & $\ast\ast\ast$ & $\ast$  & $\ast$ & $\ast\ast$ & $\ast$  & $\ast\ast$ & $\ast\ast$ & $\ast\ast\ast$ \\
\midrule
\multirow{3}{*}{AC} & Qiao & $\ast\ast\ast$ & $\ast\ast\ast$ & $\ast\ast\ast$ & $\ast\ast\ast$ & $\ast\ast\ast$ & $\ast\ast\ast$ & $\ast\ast\ast$ & $\ast\ast$ & $\ast\ast\ast$ \\
& Cilia & $\ast$ & $\ast\ast\ast$ & $\ast$ & $\ast$ & $\ast$ & $\ast$ & $\ast$ & $\ast\ast\ast$ & $\ast\ast$ \\
& Saliency4ASD & $\ast\ast\ast$ & $\ast\ast\ast$ & $\ast$ & $\ast$ & $\ast\ast\ast$ & $\ast$ & $\ast\ast\ast$ & $\ast\ast$ & $\ast\ast\ast$ \\
\midrule
\multirow{3}{*}{BIRCH} & Qiao & $\ast\ast\ast$ & $\ast\ast\ast$ & $\ast\ast\ast$ & $\ast\ast\ast$ & $\ast\ast\ast$ & $\ast\ast\ast$ & $\ast\ast\ast$ & $\ast\ast\ast$ & $\ast\ast\ast$ \\
& Cilia & $\ast$ & $\ast\ast\ast$ & $\ast$ & $\ast$ & $\ast\ast\ast$ & $\ast$ & $\ast\ast\ast$ & $\ast\ast$ & $\ast\ast$ \\
& Saliency4ASD & $\ast\ast\ast$ & $\ast\ast\ast$ & $\ast\ast$ & $\ast\ast\ast$ & $\ast\ast\ast$ & $\ast\ast$ & $\ast\ast$ & $\ast\ast$ & $\ast$ \\
\midrule
\multirow{3}{*}{DBSCAN} & Qiao & $\ast\ast\ast$ & $\ast\ast\ast$ & $\ast\ast\ast$ & $\ast\ast\ast$ & $\ast$ & $\ast\ast\ast$ & $\ast$ & $\ast\ast$ & $\ast\ast\ast$ \\
& Cilia & $\ast$ & $\ast$ & $\ast$ & $\ast\ast$ & $\ast\ast\ast$ & $\ast$ & $\ast\ast$ & $\ast\ast\ast$ & $\ast\ast\ast$ \\
& Saliency4ASD & $\ast\ast\ast$ & $\ast\ast\ast$& $\ast$ & $\ast$ & $\ast\ast\ast$ & $\ast$ & $\ast\ast\ast$ & $\ast\ast$ & $\ast$ \\
\midrule
\multirow{3}{*}{OPTICS} & Qiao & $\ast\ast\ast$ & $\ast\ast\ast$ & $\ast\ast\ast$ & $\ast\ast\ast$ & $\ast$ & $\ast\ast\ast$ & $\ast\ast\ast$ & $\ast\ast$ & $\ast\ast\ast$  \\
& Cilia & $\ast$ & $\ast\ast$ & $\ast\ast\ast$ & $\ast\ast\ast$ & $\ast\ast\ast$ & $\ast\ast\ast$ & $\ast\ast\ast$ & $\ast\ast\ast$ & $\ast\ast\ast$ \\
& Saliency4ASD & $\ast$ & $\ast\ast\ast$& $\ast\ast$ & $\ast$ & $\ast\ast\ast$ & $\ast\ast\ast$ & $\ast\ast\ast$ & $\ast$ & $\ast\ast$ \\
\midrule
\multirow{3}{*}{GMM} & Qiao & $\ast\ast\ast$ & $\ast\ast\ast$ & $\ast\ast\ast$ & $\ast\ast\ast$ & $\ast\ast\ast$ & $\ast\ast\ast$ & $\ast\ast\ast$ & $\ast$ & $\ast\ast\ast$ \\
& Cilia & $\ast$ & $\ast\ast\ast$ & $\ast$ & $\ast$ & $\ast$ & $\ast$ & $\ast$ & $\ast\ast\ast$ & $\ast\ast$ \\
& Saliency4ASD & $\ast\ast\ast$ & $\ast\ast\ast$ & $\ast$ & $\ast$ & $\ast\ast\ast$ & $\ast$ & $\ast\ast\ast$ & $\ast\ast$ & $\ast\ast$ \\
\bottomrule
\end{tabular*}
\label{tab:test}
\end{threeparttable}
\end{table*}

\section{COMPREHENSIVE BENCHMARKING OF PREDICTIVE MODELS}
Based on the first step using seven unsupervised clustering methods and the second step extracting nine different internal cluster validity indices from each method, we extracted a total of $7 \times 9$ indices features. We employed several machine learning models to utilize these 63-dimensional features to predict ASD and TD children. 

For a comprehensive benchmarking of predictive models, we selected the most suitable models from four different types of machine learning models:
\begin{itemize}
    \item \textbf{Linear Models}: Logistic Regression (LR)~\cite{nelder1972generalized} is simple, interpretable, and effective when features are linearly separable. Support Vector Machine (SVM)~\cite{hearst1998support} works well in high-dimensional spaces and can handle non-linear data using kernel functions.
    
    \item \textbf{Instance-based Models}: K-Nearest Neighbors (KNN)~\cite{peterson2009k} is simple and effective for small datasets, requiring no training phase and classifying data based on the nearest neighbors.
    
    \item \textbf{Tree-based Models}: Decision Tree provides an intuitive model for splitting data. Random Forest~\cite{breiman2001random} uses bagging to reduce overfitting, improve accuracy, and handle large datasets. XGBoost~\cite{chen2016xgboost} employs boosting, which is effective with imbalanced datasets and reduces overfitting by focusing on correcting errors from previous trees.
    
    \item \textbf{Neural Networks}: Multilayer Perceptron (MLP)~\cite{gardner1998artificial}, with a structure of $63 \times 128 \times 32 \times 1$, captures complex, non-linear relationships and is effective for large, high-dimensional data.
\end{itemize}

The performance of these predictive models is evaluated by accuracy, precision, recall, F1-score, and AUC.

\begin{table*}[t]
    \centering
    \begin{threeparttable}
    \caption{The comparison of optimal performance of predictive models across three datasets (Averaged Over 5 Runs). For each dataset, the model with the highest AUC score is marked with a \textcolor{black}{\ding{51}}. All metrics are averaged over 5 runs, each with a different random seed. \textbf{Inference Time} is represented as follows: \ding{72} for $<$0.0001s, \ding{108} for $<$0.001s, \ding{109} for $<$0.01s, and \ding{115} for $<$0.1s. Bold values represent the highest metric values for each dataset, while underlined values represent the second highest. Additionally, for each metric, the best value is \textbf{bolded}, and the second-best value is \underline{underlined}.}
    \begin{tabular*}{\textwidth}{@{\extracolsep{\fill}}llcccccc}
        \toprule
        \textbf{Dataset} & \textbf{Model} & \textbf{Accuracy} & \textbf{Precision} & \textbf{Recall} & \textbf{F1-score} & \textbf{AUC} & \textbf{Inference Time} \\
        \midrule
        \multirow{7}{*}{Qiao} 
        & LR & 0.721 & 0.713 & 0.724 & 0.715 & 0.783 & \ding{108} \\
        & SVM & 0.718 & 0.711 & 0.725 & 0.709 & 0.781 & \ding{109} \\
        & KNN & 0.702 & 0.695 & 0.692 & 0.691 & 0.742 & \ding{115} \\
        & Decision Tree & 0.669 & 0.664 & 0.668 & 0.643 & 0.713 & \ding{72} \\
        & Random Forest & 0.725 & 0.715 & 0.726 & 0.711 & 0.791 & \ding{115} \\
        & XGBoost & \underline{0.732} & \underline{0.734} & \underline{0.731} & \underline{0.732} & \underline{0.802} & \ding{108} \\
        & MLP \textcolor{black}{\ding{51}} & \textbf{0.744} & \textbf{0.743} & \textbf{0.742} & \textbf{0.741} & \textbf{0.804} & \ding{108} \\
        \midrule
        \multirow{7}{*}{Cilia} 
        & LR & 0.641 & 0.678 & 0.642 & 0.562 & 0.673 & \ding{108} \\
        & SVM & 0.623 & 0.615 & 0.621 & 0.492 & 0.682 & \ding{115} \\
        & KNN & 0.713 & 0.711 & 0.716 & \underline{0.693} & 0.746 & \ding{115} \\
        & Decision Tree & 0.684 & 0.668 & 0.679 & 0.681 & 0.675 & \ding{72} \\
        & Random Forest \textcolor{black}{\ding{51}} & \underline{0.733} & \textbf{0.742} & \textbf{0.735} & 0.701 & \textbf{0.813} & \ding{108} \\
        & XGBoost & \textbf{0.735} & \underline{0.741} & \underline{0.732} & \textbf{0.714} & \underline{0.784} & \ding{108} \\
        & MLP & 0.701 & 0.682 & 0.691 & 0.671 & 0.702 & \ding{72} \\
        \midrule
        \multirow{7}{*}{Saliency4ASD} 
        & LR & 0.713 & 0.711 & 0.713 & 0.714 & 0.753 & \ding{72} \\
        & SVM & \underline{0.732} & \underline{0.735} & \underline{0.731} & \underline{0.731} & 0.736 & \ding{108} \\
        & KNN & 0.657 & 0.655 & 0.658 & 0.656 & 0.753 & \ding{115} \\
        & Decision Tree & 0.613 & 0.605 & 0.612 & 0.601 & 0.692 & \ding{72} \\
        & Random Forest \textcolor{black}{\ding{51}} & \textbf{0.745} & \textbf{0.744} & \textbf{0.746} & \textbf{0.744} & \textbf{0.834} & \ding{109} \\
        & XGBoost & 0.712 & 0.711 & 0.712 & 0.714 & \underline{0.783} & \ding{108} \\
        & MLP & 0.681 & 0.684 & 0.683 & 0.682 & 0.771 & \ding{109} \\
        \bottomrule
    \end{tabular*}
    \label{tab:model-evaluate}
    \end{threeparttable}
\end{table*}

We conducted a benchmark experiment to explore the effectiveness of the clustering methods by employing multiple predictive models on three datasets. This experiment was conducted on the CPU with 16GB DDR4 RAM and Ubuntu 20.04 LTS system. The deep learning models were implemented using Python 3.9 with PyTorch 2.0, and traditional machine learning models were implemented using scikit-learn 1.4.0 and SciPy 1.12.0. We employed 5-fold cross-validation to ensure robust evaluation results.

Based on the average metrics across the three datasets, \textbf{Random Forest} performs the best overall, achieving the highest scores in Accuracy (0.734), Precision (0.734), Recall (0.736), F1-score (0.719), and AUC (0.813). To be specific, in the Qiao dataset, \textbf{MLP} outperformed all models, achieving the highest scores in Accuracy (0.744), Precision (0.743), Recall (0.742), F1-score (0.741), and AUC (0.804). In the Cilia dataset, \textbf{Random Forest} showed the best overall performance with the top scores in most metrics, including AUC (0.813), while \textbf{XGBoost} had a slightly higher Accuracy (0.735). For the Saliency4ASD dataset, \textbf{Random Forest} dominated across all metrics, with an AUC of 0.834, demonstrating the best performance.

These well-performed models, including MLP, Random Forest, and XGBoost, also presented high prediction efficiency. These models can all complete inference within 0.1 seconds.
\section{CONCLUSION}
We explored the use of internal cluster validity indices for analyzing gaze pattern differences between ASD and TD children. By applying seven clustering algorithms and evaluating them with nine internal cluster validity indices, we demonstrated their effectiveness in characterizing gaze patterns. Our results include a significant indices proportion of 74.1\% and an average AUC of 81\% in ASD prediction, highlighting the potential of internal cluster validity indices as valuable features for gaze analysis and ASD diagnosis. This work offers a promising direction for automating the exploration of ASD children's gaze patterns, advancing both understanding and diagnosis.

% \begin{acks}
% \end{acks}

\bibliographystyle{ACM-Reference-Format}
\bibliography{main}

\appendix

\end{document}